\documentclass[11pt,a4paper]{article}
\usepackage[dvipsnames]{xcolor}
\usepackage[hyperref]{emnlp2018}
\usepackage{times}
\usepackage{latexsym}
\usepackage[T1]{fontenc}
\usepackage{latexsym}
\usepackage{verbatim}
\usepackage{latexsym}
\usepackage{float}
\newcommand\tab[1][0.5cm]{\hspace*{#1}}
\usepackage[font={footnotesize}]{caption}
\usepackage{amsmath}
\usepackage{graphicx}
\graphicspath{ {images/} }
\usepackage{color,soul}
\usepackage{cleveref}
\usepackage{commath}
\crefformat{section}{\S#2#1#3} 
\crefformat{subsection}{\S#2#1#3}
\crefformat{subsubsection}{\S#2#1#3}
\usepackage{url}

\newcounter{Lcount}
\newcommand{\squishenum}{
\begin{list}{\arabic{Lcount}. }
{ \usecounter{Lcount}
\setlength{\itemsep}{0pt}
\setlength{\parsep}{0pt}
\setlength{\topsep}{0pt}
\setlength{\partopsep}{0pt}
\setlength{\leftmargin}{2em}
\setlength{\labelwidth}{1.5em}
\setlength{\labelsep}{0.5em} } }

\newcommand{\squishletter}{
\begin{list}{\alph{Lcount}. }
{ \usecounter{Lcount}
\setlength{\itemsep}{0pt}
\setlength{\parsep}{0pt}
\setlength{\topsep}{0pt}
\setlength{\partopsep}{0pt}
\setlength{\leftmargin}{2em}
\setlength{\labelwidth}{1.5em}
\setlength{\labelsep}{0.5em} } }

\newcommand{\squishlist}{
\begin{list}{$\bullet$}
{ \usecounter{Lcount}
\setlength{\itemsep}{0pt}
\setlength{\parsep}{0pt}
\setlength{\topsep}{0pt}
\setlength{\partopsep}{0pt}
\setlength{\leftmargin}{2em}
\setlength{\labelwidth}{1.5em}
\setlength{\labelsep}{0.5em} } }

\newcommand{\squishend}{
\end{list} }
\DeclareMathOperator*{\argmax}{arg\,max}

\newcommand{\sh}[1]{}
\newcommand{\pa}[1]{}
\newcommand{\ro}[1]{}
\newcommand{\gn}[1]{}

\usepackage{pifont}
\newcommand{\cmark}{\ding{51}}%

\aclfinalcopy 

\title{Retrieval-Based Neural Code Generation}

\author{Shirley Anugrah Hayati \tab Raphael Olivier \tab Pravalika Avvaru \\ \textbf{Pengcheng Yin\tab Anthony Tomasic\tab Graham Neubig}\\ \\
	    Language Technologies Institute, Carnegie Mellon University\\
	    {\tt \{shayati,rolivier,pavvaru,pcyin,tomasic,gneubig\}@cs.cmu.edu}
}

\date{}

\begin{document}
\maketitle
\begin{abstract}
In models to generate program source code from natural language\gn{``In models to generate program source code from natural language''}\sh{\cmark}, representing this code in a tree structure has been a common approach. However, existing methods often fail to generate complex code correctly due to a lack of ability to memorize large and complex structures\gn{Why?}\sh{\cmark}. We introduce \textsc{ReCode}, a method based on subtree retrieval that makes it possible to explicitly reference existing code examples within a neural code generation model.
First, we retrieve sentences that are similar to input sentences using a dynamic-programming-based sentence similarity scoring method. Next, we extract $n$-grams of action sequences that build the associated abstract syntax tree\gn{``abstract syntax tree''. Do not capitalize common nouns, unless you're introducing an acronym (even then it's not necessary)}\pa{\cmark}. Finally, we increase the probability of actions that cause the retrieved $n$-gram action subtree to be in the predicted code. We show that our approach improves the performance on two code generation tasks by up to +2.6 BLEU.\footnote{Code available at \url{https://github.com/sweetpeach/ReCode}} 
\end{abstract}

\section{Introduction}
Natural language to code generation, a subtask of semantic parsing, is the problem of converting natural language (NL) descriptions to code \cite{Ling2016, Yin2017, Rabinovich2017}. This task is challenging because it has a well-defined structured output and the input structure and output structure are in different forms. 

A number of neural network approaches have been proposed to solve this task. \gn{I don't think the citations in the following sentence are accurate. \citet{Vinyals2015} is not a code generation paper, and \citet{Dong2016} has some concept of structure, so it can probably be described as a tree-based model (although with weaker constraints, so it is true that it doesn't necessarily output syntactically correct code). If you want good examples of sequence-to-sequence models, \citet{Ling2016}, \citet{Jia2016}, and \citet{locascio-EtAl:2016:EMNLP2016} are better.}\ro{\cmark} Sequential approaches \cite{Ling2016,Jia2016,locascio-EtAl:2016:EMNLP2016} convert the target code into a sequence of symbols and apply a sequence-to-sequence model, but this approach does not ensure that the output will be syntactically correct. Tree-based approaches \cite{Yin2017,Rabinovich2017} represent code as Abstract Syntax Trees (ASTs), which has proven effective in improving accuracy as it enforces the well-formedness of the output code. However, representing code as a tree is not a trivial task, as the number of nodes in the tree often greatly exceeds the length of the NL description. As a result, tree-based approaches are often incapable of generating correct code for phrases in the corresponding NL description that have low frequency in the training data. 

In machine translation (MT) problems \cite{Zhang18, Gu2017, Farajian2017, Li2016}, hybrid methods combining retrieval of salient examples and neural models have proven successful in dealing with rare words. Following the intuition of these models, we hypothesize that our model can benefit from querying pairs of NL descriptions and AST structures from training data. 

In this paper, we propose \textsc{ReCode}, and adaptation of \citet{Zhang18}'s retrieval-based approach neural MT method to the code generation problem by expanding it to apply to generation of tree structures. \gn{Maybe the following two sentences are not necessary?}\pa{\cmark}Our main contribution is to introduce the use of retrieval methods in neural code generation models. We also propose a dynamic programming-based sentence-to-sentence alignment method that can be applied to similar sentences to perform word substitution and enable retrieval of imperfect matches. These contributions allow us to improve on previous state-of-the-art results. 

\section{Syntactic Code Generation}\label{codegen}
Given an NL description $q$, our purpose is to generate code (e.g. Python)\gn{by writing ``Python code'' you limit the generality of your model. Write ``code (e.g. Python)''} \pa{\cmark} represented as an AST $a$. In this work, we start with the syntactic code generation model by \citet{Yin2017}, which uses sequences of actions to generate the AST before converting it to surface code. Formally, we want to find the best generated AST $\hat{a}$ given by:
\begin{equation*}
\vspace{-7pt}
\hat{a} = \argmax_{a} p(a|q)
\end{equation*}
\vspace{-5pt}
\begin{equation*}
\vspace{-5pt}
p(a|q) = \prod_{t=1}^{T} p(y_t|y_{<t}, q)
\end{equation*}
where $y_t$ is the action taken at time step $t$ and $y_{<t} = y_1 ... y_{t-1}$ and $T$ is the number of total time steps of the whole action sequence resulting in AST $a$. \gn{The relationship between $a$ and $y$ is not specified here.}\ro{\cmark}

We have two types of actions to build an AST: \textsc{ApplyRule} and \textsc{GenToken}. \textsc{ApplyRule(}\textit{r}\textsc{)} \gn{the function arguments should not be bold. also, in the original paper ApplyRule and GenToken are ``textsc'' not ``texttt/textbf''. I prefer this if you don't have any reason to do otherwise.} \pa{\cmark} expands the current node in the tree by applying production rule \textit{r} from the abstract syntax grammar\gn{``from the abstract syntax grammar''}\pa{\cmark}\footnote{\url{https://docs.python.org/2/library/ast.html}} to the current node. \textsc{GenToken(}\textit{v}\textsc{)} populates terminal nodes with the variable \textit{v} which can be generated from vocabulary or by \textsc{Copy}ing variable names or values from the NL description. The generation process follows a preorder traversal starting with the \texttt{root} node. Figure \ref{fig:ASTAction} shows an action tree for the example code: the nodes correspond to actions per time step in the construction of the AST.

Interested readers can reference \citet{Yin2017} for more detail of the neural model, which consists of a bidirectional LSTM \cite{Hochreiter97} encoder-decoder with action embeddings, context vectors, parent feeding, and a copy mechanism using pointer networks. 

\section{\textsc{ReCode}: Retrieval-Based Neural Code Generation}\label{retrieval}
We propose \textsc{ReCode}, a method for retrieval-based neural syntactic code generation, using retrieved action subtrees. Following \citet{Zhang18}'s method for neural machine translation, these retrieved subtrees act as templates that bias the generation of output code. Our pipeline at test time is as follows:
\squishlist
\item retrieve from the training set NL descriptions that are most similar with our input sentence (\cref{retrieval}),
\item extract \textbf{$\mathbf{n}$-gram action subtrees} from these retrieved sentences' corresponding target ASTs (\cref{ngrams}),
\item alter the copying actions in these subtrees, by substituting words of the retrieved sentence with corresponding words in the input sentence (\cref{copy_actions}), and
\item at every decoding step, increase the probability of actions that would lead to having these subtrees in the produced tree (\cref{retrievalcodegen}).
\squishend

\subsection{Retrieval of Training Instances}
\label{retrieval}
For every retrieved NL description $q_m$ from training set (or \textit{retrieved sentence} for short), we compute its similarity with input $q$, using a sentence similarity formula \cite{Gu2016,Zhang18}:
$$
\text{sim}(q,q_m) = 1 - \frac{d(q,q_m)}{\text{max}(\abs{q},\abs{q_m})}
$$
where $d$ is the edit distance.
\gn{I don't see $d(X,X_m)$ defined anywhere. Could you write one sentence about exactly what formula you use?}\ro{\cmark}
\gn{For multi word function names, it's better to use \texttt{``text\{\}''}. I fixed it here, please try to fix the rest.}
We retrieve only the top $M$ sentences according to this metric where $M$ is a hyperparameter. These scores will later be used to increase action probabilities accordingly.

\subsection{Extracting $N$-gram Action Subtrees}
\label{ngrams}
In \citet{Zhang18}, they collect $n$-grams from the output side of the retrieved sentences and encourage the model to generate these $n$-grams. Word $n$-grams are obvious candidates when generating a sequence of words as output, as in NMT. However, in syntax-based code generation, the generation target is ASTs with no obvious linear structure. To resolve this problem, we instead use \emph{retrieved pieces} of $n$-gram subtrees from the target code corresponding to the retrieved NL descriptions. Though we could select successive nodes in the AST as retrieved pieces, such as {\small \texttt{[assign; expr*(targets); expr]}} from Figure \ref{fig:ASTAction}, we would miss important structural information from the rules that are used. Thus, we choose to exploit actions in the generation model rather than AST nodes themselves to be candidates for our retrieved pieces.

In the action tree (\hyperref[fig:ASTAction]{Figure 1}), we considered only successive actions, such as subtrees where each node has one or no children, to avoid overly rigid structures or combinatorial explosion of the number of retrieved pieces the model has to consider. For example, such an action subtree would be given by {\small \texttt{[assign $\rightarrow$ expr*(targets), expr(value) ; expr(value) $\rightarrow$ List; List $\rightarrow$ epsilon]}}.

As the node in the action tree holds structural information about its children, we set the subtrees to have a fixed depth, linear in the size of the tree. These can be considered ``$n$-grams of actions'', emphasizing the comparison with machine translation which uses $n$-grams of words. $n$ is a hyperparameter to be tuned. 


\subsection{Word Substitution in Copy Actions}
\label{copy_actions}
\begin{figure}
\centering\includegraphics[scale=0.35]{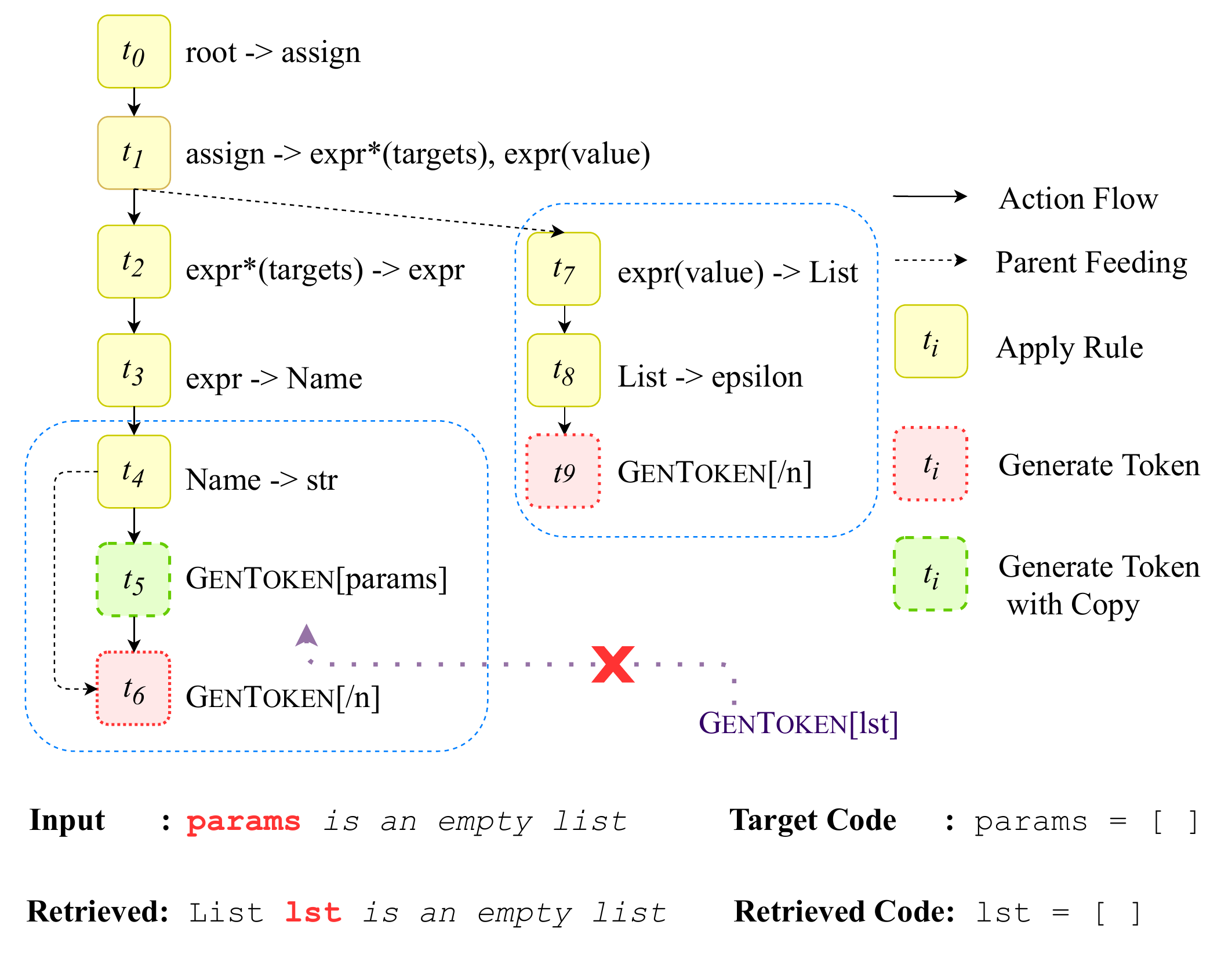}
\caption{The action sequence used to generate AST for the target code given the input example. Dashed nodes represent terminals. Each node is labeled with time steps. \textsc{ApplyRule} action is represented as rule in this figure. Blue dotted boxes denote 3-gram action subtrees.  Italic words are unedited words. Red bold words are different object names. \sh{I updated the figure following the reviewers' comment}} 
\label{fig:ASTAction}
\vspace{-5mm}
\end{figure} 
Using the retrieved subtree without modification is problematic if it contains at least one node corresponding to a \textsc{Copy} action because copied tokens from the retrieved sentence may be different from those in the input.  Figure \ref{fig:ASTAction} shows an example when the input and retrieved sentence have four common words, but the object names are different. The extracted action $n$-gram would contain the rule that copies the second word (``\textcolor{WildStrawberry}{lst}'') of the retrieved sentence while we want to copy the first word (``\textcolor{WildStrawberry}{params}'') from the input.


By computing word-based edit distance between the input description and the retrieved sentence, we implement a one-to-one sentence alignment method that infers correspondences between uncommon words. For unaligned words, we alter all \textsc{Copy} rules in the extracted $n$-grams to copy tokens by their aligned counterpart, such as replace ``\textcolor{WildStrawberry}{params}'' with ``\textcolor{WildStrawberry}{lst}'', and delete the $n$-gram subtree, as it is not likely to be relevant in the predicted tree. Thus, in the example in Figure \ref{fig:ASTAction}, the \textsc{GenToken(lst)} action in $t_5$ will not be executed. 


\subsection{Retrieval-Guided Code Generation}\label{retrievalcodegen}
$N$-gram subtrees from all retrieved sentences are assigned a score, based on the best similarity score of all instances where they appeared. We normalize the scores for each input sentence by subtracting the average over the training dataset.

At decoding time, incorporate these retrieval-derived scores into beam search: for a given time step, all actions that would result in one of the retrieved $n$-grams $u$ to be in the prediction tree has its log probability  $\mathrm{log} (p(y_t\mid y_1^{t-1}))$ increased by $
\lambda*\text{score}(u)
$
where $\lambda$ is a hyperparameter, and $\text{score}(u)$ is the maximal $\text{sim}(q, q_m)$ from which $u$ is extracted. The probability distribution is then renormalized.

\section{Datasets and Evaluation Metrics}
\vspace{-3mm}
\begin{table}[t]
\centering
\small
\begin{tabular}{l|ll}
\hline
\textbf{Dataset}     & \textbf{HS} & \textbf{Django}\\
\hline
Train & 533 & 16,000\\
Dev     & 66 & 1,000\\
Test & 66 & 1,805\\ \hline
Avg. tokens in description & 39.1 & 14.3\\
Avg. number of nodes of AST & 136.6 & 17.2\\
\hline
\end{tabular}
\caption{Dataset statistics as reported \citet{Yin2017} \gn{I moved this to the top of the page, as specified by the ACL style}}
\label{tab:datasets}
\vspace{-3mm}
\end{table}

We evaluate \textsc{ReCode} with the Hearthstone (HS) \cite{Ling2016} and Django \cite{Oda2015} datasets, as preprocessed by \citet{Yin2017}. HS consists of Python classes that implement Hearthstone card descriptions while Django contains pairs of Python source code and English pseudo-code from Django web framework. Table \ref{tab:datasets} summarizes dataset statistics.

For evaluation metrics, we use accuracy of exact match and the BLEU score following \citet{Yin2017}.

\section{Experiments}\label{experiments}
For the neural code generation model, we use the settings explained in \citet{Yin2017}. For the retrieval method, we tuned hyperparameters and achieved best result when we set  $n_{max} = 4$ and $\lambda = 3$ for both datasets\footnote{$n$-gram subtrees are collected up to $n_{max}$-gram}. For HS, we set $M = 3$ and $M = 10$ for Django. 

We compare our model with \citet{Yin2017}'s model that we call \textsc{YN17} for brevity, and a sequence-to-sequence (\textsc{Seq2Seq}) model that we implemented. \textsc{Seq2Seq} is an attention-enabled encoder-decoder model \cite{Bahdanau2014}. The encoder is a bidirectional LSTM and the decoder is an LSTM. \sh{I explained more about \textsc{Seq2Seq} following the reviewers' feedback}

\subsection{Results}
Table \ref{mainresults} shows that \textsc{ReCode} outperforms the baselines in both BLEU and accuracy,
providing evidence for the effectiveness of incorporating retrieval methods into tree-based approaches. 

\begin{table}[H]
\centering
\small
\begin{tabular}{lrr|rr}
\hline & \multicolumn{2}{c}{\textbf{HS}}                                      & \multicolumn{2}{c}{\textbf{Django}}\\ 
 & \multicolumn{1}{c}{\textbf{Acc}} & \multicolumn{1}{c}{\textbf{BLEU}} & \textbf{Acc}          & \textbf{BLEU}         \\ \hline
\textsc{Seq2Seq}  & 0.0  & 55.0  & 13.9 & 67.3                 \\
\textsc{YN17}     & 16.2                             & 75.8   & 71.6                  & 84.5                  \\
ASN$^\dagger$ & 18.2  & 77.6 & \multicolumn{1}{c}{-} & \multicolumn{1}{c}{-} \\ 
ASN + \textsc{SupAtt}$^\dagger$ & \textit{22.7}                   & \textit{79.2} & \multicolumn{1}{c}{-} & \multicolumn{1}{c}{-} \\
\hline
\textsc{ReCode}   & \textbf{19.6} & \textbf{78.4} & \textbf{72.8} & \textbf{84.7}         \\ \hline
\end{tabular}
\caption{Results compared to baselines. YN17 result is taken from \citet{Yin2017}. ASN result is taken from \citet{Rabinovich2017}}. 
\label{mainresults}
\vspace{-3mm}
\end{table}

\sh{I added the statistical significance test result} We ran statistical significance tests for \textsc{ReCode} and YN17, using bootstrap resampling with $N$ = 10,000. For the BLEU scores of both datasets, $p < 0.001$. For the exact match accuracy, $p < 0.001$ for Django dataset, but for Hearthstone, $p > 0.3$, showing that the retrieval-based model is on par with YN17. It is worth noting, though,  that HS consists of long and complex code, and that generating exact matches is very difficult, making exact match accuracy a less reliable metric. 

We also compare \textsc{ReCode} with \citet{Rabinovich2017}'s Abstract Syntax Networks with supervision (ASN+\textsc{SupAtt}) which is the state-of-the-art system for HS. \sh{The reviewers would like to have ASN+\textsc{SupAtt} to be included in the table.} \textsc{ReCode} exceeds ASN without extra supervision though ASN+\textsc{SupAtt} has a slightly better result. However, ASN+\textsc{SupAtt} is trained with supervised attention extracted through heuristic exact word matches while our attention is unsupervised. 


\subsection{Discussion and Analysis}
From our observation and as mentioned in \citet{Rabinovich2017}, HS contains classes with similar structure, so the code generation task could be simply matching the tree structure and filling the terminal tokens with correct variables and values. However, when the code consists of complex logic, partial implementation errors occur, leading to low exact match accuracy \cite{Yin2017}. Analyzing our result, we find this intuition to be true not only for HS but also for Django. 

\begin{table}
\centering
\tiny
\begin{tabular}{lr}
\hline
Example 1
\\
\hline
``{\fontfamily{lmss}\selectfont if offset is lesser than integer 0, sign is set to '-', otherwise sign is '+' ''}& \textbf{Input} \\
\texttt{sign \textcolor{WildStrawberry}{\textbf{=}} offset \textcolor{WildStrawberry}{\textbf{<}} \textcolor{Plum}{\textbf{0}} \textcolor{WildStrawberry}{\textbf{or}} \textcolor{Peach}{\textbf{'-'}}} & \textbf{YN17}\\
\texttt{sign \textcolor{WildStrawberry}{\textbf{=}} \textcolor{Peach}{\textbf{'-'}} \textcolor{WildStrawberry}{\textbf{if}} offset \textcolor{WildStrawberry}{\textbf{<}} \textcolor{Plum}{\textbf{0}} \textcolor{WildStrawberry}{\textbf{else}} \textcolor{Peach}{\textbf{'+'}}} & \textbf{\textsc{ReCode}}\\
\texttt{sign \textcolor{WildStrawberry}{\textbf{=}} \textcolor{Peach}{\textbf{'-'}} \textcolor{WildStrawberry}{\textbf{if}} offset \textcolor{WildStrawberry}{\textbf{<}} \textcolor{Plum}{\textbf{0}} \textcolor{WildStrawberry}{\textbf{else}} \textcolor{Peach}{\textbf{'+'}}} & \textbf{Gold}
\\
\hline
Example 2
\\
\hline
``{\fontfamily{lmss}\selectfont evaluate the function timesince with d, now and reversed set} & \textbf{Input}\\
{\fontfamily{lmss}\selectfont to boolean true as arguments, return the result.''}\\
\texttt{\textcolor{WildStrawberry}{\textbf{return}} \textcolor{Cyan}{\textbf{reversed}}(d, \textcolor{Orange}{\textbf{reversed}}\textcolor{WildStrawberry}{\textbf{=}}now)} & \textbf{YN17}\\
\texttt{\textcolor{WildStrawberry}{\textbf{return}} \textcolor{Cyan}{\textbf{timesince}}(d, now, \textcolor{Orange}{\textbf{reversed}}\textcolor{WildStrawberry}{\textbf{=}}now)} & \textbf{\textsc{ReCode}}\\
\texttt{\textcolor{WildStrawberry}{\textbf{return}} \textcolor{Cyan}{\textbf{timesince}}(d, now, \textcolor{Orange}{\textbf{reversed}}\textcolor{WildStrawberry}{\textbf{=}}True)} & \textbf{Gold}
\\
\hline
Example 3
\\
\hline
``{\fontfamily{lmss}\selectfont return an instance of SafeText , } & \textbf{Input}\\
{\fontfamily{lmss}\selectfont created with an argument s converted into a string .}''\\
%
\texttt{\textcolor{WildStrawberry}{\textbf{return}} \textcolor{Cyan}{\textbf{SafeText}}(\textcolor{Cyan}{\textbf{\textit{bool}}}(s))} & \textbf{YN17}\\
\texttt{\textcolor{WildStrawberry}{\textbf{return}} \textcolor{Cyan}{\textbf{SafeText}}(s)} & \textbf{\textsc{ReCode}}\\
\texttt{\textcolor{WildStrawberry}{\textbf{return}} \textcolor{Cyan}{\textbf{SafeString}}(\textcolor{Cyan}{\textbf{\textit{str}}}(s))} & \textbf{Gold}
\\
\hline
\end{tabular}
\caption{Django examples on correct code and predicted code with retrieval (RECODE) and without retrieval (YN17).}
\label{tab:django_code_example}
\vspace{-5mm}
\end{table}

Examining the generated output for the Django dataset in Table \ref{tab:django_code_example}, we can see that in the first example, our retrieval model can successfully generate the correct code when \textsc{YN17} fails. This difference suggests that our retrieval model benefits from the action subtrees from the retrieved sentences. In the second example, although our generated code does not perfectly match the reference code, it has a higher BLEU score compared to the output of \textsc{YN17} because our model can predict part of the code (\texttt{timesince(d, now, reversed)}) correctly. The third example shows where our method fails to apply the correct action as it cannot cast \texttt{s} to \texttt{str} type while \textsc{YN17} can at least cast \texttt{s} into a type (\texttt{bool}). Another common type of error that we found \textsc{ReCode}'s generated outputs is incorrect variable copying, similarly to what is discussed in \citet{Yin2017} and \citet{Rabinovich2017}. 

Table \ref{tab:hs_code_example} presents a result on the HS dataset\footnote{More example of HS code is provided in the supplementary material.}. We can see that our retrieval model can handle complex code more effectively. \pa{removed the following line} 

\begin{table}
\centering
\tiny
\begin{tabular}{l}
\hline
{\fontfamily{lmss}\selectfont NAME\_BEGIN Earth Elemental NAME\_END ATK\_BEGIN 7} \tab {\textbf{Input}}\\
{\fontfamily{lmss}\selectfont ATK\_END DEF\_BEGIN 8 DEF\_END COST\_BEGIN 5}\\
{\fontfamily{lmss}\selectfont COST\_END DUR\_BEGIN -1 DUR\_END TYPE\_BEGIN Minion }\\
{\fontfamily{lmss}\selectfont TYPE\_END PLAYER\_CLS\_BEGIN Shaman PLAYER\_CLS\_END}\\
{\fontfamily{lmss}\selectfont RACE\_BEGIN NIL RACE\_END RARITY\_BEGIN  Epic RARITY\_END}\\
{\fontfamily{lmss}\selectfont DESC\_BEGIN Taunt . Overload : ( 3 ) DESC\_END.}
\\
\hline
\texttt{\textcolor{Cyan}{\textbf{class}} \textcolor{Green}{\textbf{EarthElemental}} (\textcolor{Green}{\textbf{\textit{MinionCard}}}) :} \tab \tab \tab \tab {\textbf{YN17}}\\
\texttt{\tab \textcolor{Cyan}{\textbf{def \_\_init\_\_}} (\textcolor{Orange}{\textbf{\textit{self}}}) :}\\
\texttt{  \tab \tab \textcolor{Cyan}{\textbf{super}} ( ).\textcolor{Cyan}{\textbf{\_\_init\_\_}} ("Earth Elemental", \textcolor{Plum}{\textbf{5}},}\\
\texttt{  \tab \tab \tab[0.1cm] CHARACTER\_CLASS.SHAMAN, CARD\_RARITY.EPIC, }\\
\texttt{  \tab \tab \tab[0.1cm] \textcolor{Orange}{\textbf{\textit{buffs}}}=[\textcolor{Cyan}{\textbf{Buff}}(\textcolor{Cyan}{\textbf{ManaChange(Count}}}\\
\texttt{\tab \tab \tab[0.1cm](\textcolor{Cyan}{\textbf{MinionSelector}}(None, \textcolor{Cyan}{\textbf{BothPlayer}}())), \textcolor{WildStrawberry}{\textbf{-}}\textcolor{Plum}{\textbf{1}}))])}\\
\texttt{\tab  \textcolor{Cyan}{\textbf{def}} \textcolor{Green}{\textbf{create\_minion}} (\textcolor{Orange}{\textbf{\textit{self}}}, \textcolor{Orange}{\textbf{\textit{player}}}) :}\\ 
\texttt{ \tab \tab return \textcolor{Cyan}{\textbf{Minion}}(\textcolor{Plum}{\textbf{7}}, \textcolor{Plum}{\textbf{8}}, \textcolor{Orange}{\textbf{\textit{taunt}}}\textcolor{WildStrawberry}{\textbf{=}}\textcolor{Plum}{\textbf{True}}) } \\
\newline
\\
\hline
\texttt{\textcolor{Cyan}{\textbf{class}} \textcolor{Green}{\textbf{EarthElemental}} (\textcolor{Green}{\textbf{\textit{MinionCard}}}) :} \tab \tab \tab \tab {\textbf{\textsc{ReCode}}}\\
\texttt{\tab \textcolor{Cyan}{\textbf{def \_\_init\_\_}} (\textcolor{Orange}{\textbf{\textit{self}}}) :}\\
\texttt{  \tab \tab \textcolor{Cyan}{\textbf{super}} ( ).\textcolor{Cyan}{\textbf{\_\_init\_\_}} ("Earth Elemental", \textcolor{Plum}{\textbf{5}},}\\
\texttt{  \tab \tab \tab[0.1cm] CHARACTER\_CLASS.SHAMAN, CARD\_RARITY.EPIC, }\\
\texttt{  \tab \tab \tab[0.1cm] \textcolor{Orange}{\textbf{\textit{overload}}}\textcolor{WildStrawberry}{\textbf{=}}\textcolor{Plum}{\textbf{3}})}\\
\texttt{\tab  \textcolor{Cyan}{\textbf{def}} \textcolor{Green}{\textbf{create\_minion}} (\textcolor{Orange}{\textbf{\textit{self}}}, \textcolor{Orange}{\textbf{\textit{player}}}) :}\\ 
\texttt{ \tab \tab return \textcolor{Cyan}{\textbf{Minion}}(\textcolor{Plum}{\textbf{7}}, \textcolor{Plum}{\textbf{8}}, \textcolor{Orange}{\textbf{\textit{taunt}}}\textcolor{WildStrawberry}{\textbf{=}}\textcolor{Plum}{\textbf{True}}) } \\
\newline
\\
\hline
\texttt{\textcolor{Cyan}{\textbf{class}} \textcolor{Green}{\textbf{EarthElemental}} (\textcolor{Green}{\textbf{\textit{MinionCard}}}) :} \tab \tab \tab \tab {\textbf{Gold}}\\
\texttt{\tab \textcolor{Cyan}{\textbf{def \_\_init\_\_}} (\textcolor{Orange}{\textbf{\textit{self}}}) :}\\
\texttt{  \tab \tab \textcolor{Cyan}{\textbf{super}} ( ).\textcolor{Cyan}{\textbf{\_\_init\_\_}} ("Earth Elemental", \textcolor{Plum}{\textbf{5}},}\\
\texttt{  \tab \tab \tab[0.1cm] CHARACTER\_CLASS.SHAMAN, CARD\_RARITY.EPIC, }\\
\texttt{  \tab \tab \tab[0.1cm] \textcolor{Orange}{\textbf{\textit{overload}}}\textcolor{WildStrawberry}{\textbf{=}}\textcolor{Plum}{\textbf{1}})}\\
\texttt{\tab  \textcolor{Cyan}{\textbf{def}} \textcolor{Green}{\textbf{create\_minion}} (\textcolor{Orange}{\textbf{\textit{self}}}, \textcolor{Orange}{\textbf{\textit{player}}}) :}\\ 
\texttt{ \tab \tab return \textcolor{Cyan}{\textbf{Minion}}(\textcolor{Plum}{\textbf{7}}, \textcolor{Plum}{\textbf{8}}, \textcolor{Orange}{\textbf{\textit{taunt}}}\textcolor{WildStrawberry}{\textbf{=}}True) } 
\newline
\\
\hline
\\
\end{tabular}
\caption{HS examples on correct code and predicted code with retrieval (\textsc{ReCode}) and without retrieval (YN17).} 
\label{tab:hs_code_example}
\vspace{-5mm}
\end{table}

\section{Related Work}

\gn{Cite work on subtree-based parsing models, e.g. \cite{bod1992computational,galley-EtAl:2006:COLACL}} \ro{\cmark in the end}

Several works on code generation focus on domain specific languages \cite{Raza2015,Kushman2013}. For general purpose code generation, some data-driven work has been done for predicting input parsers \cite{Lei2013} or a set of relevant methods \cite{Raghothaman2015}. Some attempts using neural networks have used sequence-to-sequence models \cite{Ling2016} or tree-based architectures \cite{Dong2016,Alvarez2017}. \citet{Ling2016,Jia2016,locascio-EtAl:2016:EMNLP2016} treat semantic parsing as a sequence generation task by linearizing trees. The closest work to ours are \citet{Yin2017} and \citet{Rabinovich2017} which represent code as an AST. 
Another close work is \citet{dong-lapata:2018:Long}, which uses a two-staged structure-aware neural architecture. They initially generate a low-level sketch and then fill in the missing information using the NL and the sketch.  
\gn{Cite this contemporaneous work: \cite{dong-lapata:2018:Long}.}\pa{\cmark}

Recent works on retrieval-guided neural machine translation have been presented by \citet{Gu2017,Farajian2017,Li2016,Zhang18}. \citet{Gu2017} use the retrieved sentence pairs as extra inputs to the NMT model. \citet{Zhang18} employ a simpler and faster retrieval method to guide neural MT where translation pieces are $n$-grams from retrieved target sentences. We modify \citet{Zhang18}'s method from textual $n$-grams to $n$-grams over subtrees to exploit the code structural similarity, and propose methods to deal with complex statements and rare words. 


In addition, some previous works have used subtrees in structured prediction tasks. For example, \citet{galley-EtAl:2006:COLACL} used them in syntax-based translation models. In \citet{galley-EtAl:2006:COLACL}, subtrees of the input sentence's parse tree are associated with corresponding words in the output sentence. 

\section{Conclusion}
We proposed an action subtree retrieval method at test time on top of an AST-driven neural model for generating general-purpose code. The predicted surface code is syntactically correct, and the retrieval component improves the performance of a previously state-of-the-art model\gn{``unsupervised-attention'' doesn't make much sense to me here...}\sh{\cmark}. Our successful result suggests that our idea of retrieval-based generation can be potentially applied to other tree-structured prediction tasks. 

\section*{Acknowledgements}

\gn{If anyone else who is not a co-author gave you some useful advice, feel free to thank them here.}

\gn{If you have received any other funding for your stipend, tuition, etc., please also add acknowledgement of it here.} 
We are grateful to Lucile Callebert for insightful discussions, Aldrian Obaja Muis for helpful input on early version writing, and anonymous reviewers for useful feedback.\pa{\cmark} 
This material is based upon work supported by the National Science Foundation under Grant No. 1815287.

\bibliography{emnlp2018}
\bibliographystyle{acl_natbib_nourl}

\end{document}